\UseRawInputEncoding

\documentclass[sigconf]{acmart} 
\AtBeginDocument{%
  \providecommand\BibTeX{{%
    \normalfont B\kern-0.5em{\scshape i\kern-0.25em b}\kern-0.8em\TeX}}}

\setcopyright{acmcopyright}
\copyrightyear{2021}
\acmYear{2021}

\usepackage{makecell}
\usepackage{url}
\usepackage{multicol}
\usepackage{graphicx}
\usepackage{xcolor}

\begin{document}

\title[Deep Fusion Models for SARS-CoV-2 Inhibitor Screening]{High-Throughput Virtual Screening of Small Molecule Inhibitors for SARS-CoV-2 Protein Targets with Deep Fusion Models}

\author{Garrett A. Stevenson$^1$, Derek Jones$^1$, Hyojin Kim$^1$, W. F. Drew Bennett$^1$, Brian J. Bennion$^1$, Monica Borucki$^1$, Feliza Bourguet$^1$, Aidan Epstein$^1$, Magdalena Franco$^1$, Brooke Harmon$^2$, Stewart He$^1$, Max P. Katz$^3$, Daniel Kirshner$^1$, Victoria Lao$^1$, Edmond Y. Lau$^1$, Jacky Lo$^1$, Kevin McLoughlin$^1$, Richard Mosesso$^2$, Deepa K. Murugesh$^1$, Oscar A. Negrete$^2$, Edwin A. Saada$^1$, Brent Segelke$^1$, Maxwell Stefan$^2$, Marisa W. Torres$^1$, Dina Weilhammer$^1$, Sergio Wong$^1$, Yue Yang$^1$, Adam Zemla$^1$, Xiaohua Zhang$^1$, Fangqiang Zhu$^1$, Felice C. Lightstone$^1$, Jonathan E. Allen$^1$}
\email{allen99@llnl.gov}
\authornotemark[1]
\affiliation{%
  \institution{$^1$Lawrence Livermore National Laboratory}
  \streetaddress{7011 East Ave}
  \city{Livermore}
  \state{California}
  \country{USA}
  \postcode{94550}
}
\affiliation{%
  \institution{$^2$Sandia National Laboratories}
  \streetaddress{7000 East Ave}
  \city{Livermore}
  \state{California}
  \country{USA}
  \postcode{94550}
}
\affiliation{%
  \institution{$^3$NVIDIA Corporation}
  \streetaddress{2788 San Tomas Expy}
  \city{Santa Clara}
  \state{California}
  \country{USA}
  \postcode{95051}
}
\renewcommand{\shortauthors}{G. A. Stevenson, et al.}

\begin{abstract}
Structure-based Deep Fusion models were recently shown to outperform several physics- and machine learning-based protein-ligand binding affinity prediction methods. As part of a multi-institutional COVID-19 pandemic response, over 500 million small molecules were computationally screened against four protein structures from the novel coronavirus (SARS-CoV-2), which causes COVID-19. Three enhancements to Deep Fusion were made in order to evaluate more than 5 billion docked poses on SARS-CoV-2 protein targets. First, the Deep Fusion concept was refined by formulating the architecture as one, coherently backpropagated model (Coherent Fusion) to improve binding-affinity prediction accuracy. Secondly, the model was trained using a distributed, genetic hyper-parameter optimization. Finally, a scalable, high-throughput screening capability was developed to maximize the number of ligands evaluated and expedite the path to experimental evaluation. In this work, we present both the methods developed for machine learning-based high-throughput screening and results from using our computational pipeline to find SARS-CoV-2 inhibitors. 
\end{abstract}




\maketitle

\section{Introduction}
The COVID-19 disease caused by the severe acute respiratory syndrome coronavirus (SARS-CoV-2) is responsible for the most recent, severe pandemic in modern human history \cite{lauetal}. At the onset of the COVID-19 pandemic, a worldwide effort began to identify and provide target proteins for vaccine and drug development to neutralize the virus. Two distinct proteins were rapidly solved; the trimeric spike protein (spike), which binds to human ACE2 to enter human cells  \cite{belouzard2012mechanisms} and the main protease (M\textsuperscript{pro}), which plays a pivotal role in viral gene expression and replication \cite{ullrich2020sars}. In response to the pandemic, we participated in a large-scale multi-institutional effort to virtually screen, experimentally test, and optimize therapeutic leads targeting the spike and M\textsuperscript{pro} SARS-CoV-2 protein targets. Two different binding sites from the spike protein (denoted spike1, spike2) and two different conformations of the M\textsuperscript{pro} active site (denoted protease1, protease2) were used in the high-throughput screening calculations.

Experimental tests of drug candidates are expensive and serve as a fundamental bottleneck in drug discovery. As a result, computational chemistry is used extensively to accelerate the discovery process by screening drugs and nominating the strongest candidates for experimental validation \cite{wongetal}. High Performance Computing (HPC) plays a critical role in virtual screening \cite{zhang2013message, zhang2014toward,jorgensen2004many, kitchen2004docking, cheng2012structure} by accelerating computationally expensive calculations and providing the scalability necessary to screen large numbers of candidate molecules. This is crucial, as the chemical space of potential molecules has been estimated to be on the order of 10\textsuperscript{60} \cite{reymond2010chemical, drew2012size}. Accurately estimating protein-ligand binding affinities is an important step in drug discovery. However, even computationally expensive, biophysics-based scoring methods find predicting binding free energy a difficult task \cite{wongetal, jones2020improved, doi:10.1021/acs.jcim.0c00026}. Deep learning methods represent an alternative, rapid approach to binding affinity prediction which alleviates the dependence on hand-curated features, which may not capture the mechanism of binding \cite{ballesteretal, ain2015machine}. 

The two leading deep learning approaches to structure-based binding affinity prediction fall into two categories: 3-dimensional Convolutional Neural Networks (3D-CNNs) \cite{ragoza2017protein,jimenez2018k,zhang2019deepbindrg} and Spatial Graph Convolutional Neural Networks (SG-CNNs) \cite{feinberg2018potentialnet,zhou2020distance, lim2019predicting}. Fundamentally, 3D-CNN models exploit a voxelized representation of atoms in a 3D grid, which portrays protein-ligand compounds in a Euclidean space for inference. On the other hand, SG-CNN approaches leverage a graph representation of the protein-ligand complex, allowing for multiple “edge-types” to be encoded in the representation (\textit{e.g.,} distinct distance thresholds corresponding to covalent or non-covalent interactions) to sub-select groups of atoms and evaluate their pairwise interactions.

These methods are significantly different in the way they represent compounds and their mechanisms of inference. However, they seek to achieve the same goal: accurate prediction of binding free energy on novel compounds. This observation led to the hypothesis that the 3D-CNN and SG-CNN likely have complementary strengths, which could be exploited by fusing the latent spaces of each model’s learned features . This approach to “Fusion” modeling was explored and shown to achieve superior generalization performance in predicting protein-ligand binding on X-ray crystallographic structures and virtually-docked poses of protein-ligand compounds in hold-out test sets \cite{jones2020improved}. In this work, we detail improvements to Fusion, describe its utility in high-throughput screening, and evaluate its application to the SARS-CoV-2 drug discovery problem.  A key innovation reported here is the potential to train a fusion model "coherently”. While Coherent Fusion comes with the increased computational burden of training a more complex model, the increase in complexity is mitigated by using HPC to perform parallel, distributed training.
\section{Deep Fusion}
\subsection{Fusion Modeling in Computational Chemistry}
Machine learning, specifically deep learning, approaches to protein-ligand binding affinity prediction, represent a promising new development in drug discovery \cite{doi:10.1021/acs.jcim.0c00026, ballesteretal, ain2015machine, jimenez2018k, zhang2019deepbindrg, feinberg2018potentialnet, zhou2020distance, gomes2017atomic, stepniewska2017pafnucy}. At a high level, the deep-learned models being proposed for binding affinity prediction are single-pass, feed-forward systems. This fundamental model formulation results in a computational advantage, in that, the models quickly predict binding affinity in one pass over their input. The simplicity and speed of deep learning prediction relative to biophysics-based computations make them especially attractive in the context of massive virtual drug screens.

The concept of fusion is a recent development in deep learning, initially applied to computer vision problems \cite{roitberg2019analysis, yang2018fusion, wagner2016multispectral, li2017multimodal}. In fusion, models are combined by integrating different modes of data or approaches for more predictive power. This concept was recently applied to computational chemistry in the form of Fusion models for Atomic and molecular STructures (FAST) \cite{jones2020improved}, where fusion of the two leading deep learning models (3D-CNN, SG-CNN) was shown to improve binding affinity prediction. Specifically, the Late Fusion and Mid-level Fusion models are shown as approximately equivalent or superior to individual SG-CNNs, individual 3D-CNNs, other state-of-the-art deep learning models \cite{jimenez2018k,stepniewska2017pafnucy}, and physics-based approaches including both Autodock Vina \cite{trott2010autodock} and Molecular Mechanics - Generalized Born / Surface Area (MM/GBSA) which has been shown to improve ligand pose ranking for certain target proteins \cite{wongetal, hou2011assessing}. The Late Fusion and Mid-level Fusion FAST models were specifically described and made publicly available. In the context of this work, we retrain and optimize the models on data from PDBbind-2019 \cite{wang2005PDBbind}.
\begin{equation}
pK = -log(K);\: where\: K=K_{i},K_{d}, or\:IC_{50}\label{eq:1}
\end{equation}
The Late Fusion approach is simple, but its superior performance compared to individual SG-CNNs and 3D-CNNs shows the potential of fusion modeling. In Late Fusion, SG-CNN and 3D-CNN models were separately trained to predict absolute binding affinity (Equation \ref{eq:1}), where absolute binding affinity is defined as the negative logarithm of a binding constant $K$. Binding affinity data in this study is measured as an inhibitory constant $K_{i}$, disassociation constant $K_{d}$. or inhibitory activity ($IC_{50}$), where these measurements are treated as equivalent labels in our calculations. Late Fusion takes the unweighted arithmetic mean across the individually-predicted binding affinity values from the SG-CNN and 3D-CNN models.

The Mid-level Fusion model is also described in \cite{jones2020improved}. The model is defined by extracting the latent space feature vector from $Layer^{N-3}$ of an N-layer SG-CNN and $Layer^{M-1}$ of an M-layer 3D-CNN. Each vector of gradients is then processed through model-specific dense layers, concatenated with the originally extracted vectors, and passed through two fusion dense layers for a final prediction. In contrast to Late Fusion, Mid-level Fusion is a non-linear combination of the respective models and its performance has been shown to outperform Late Fusion in some cases.
\begin{figure*}
  \centering
  \includegraphics[width=\textwidth]{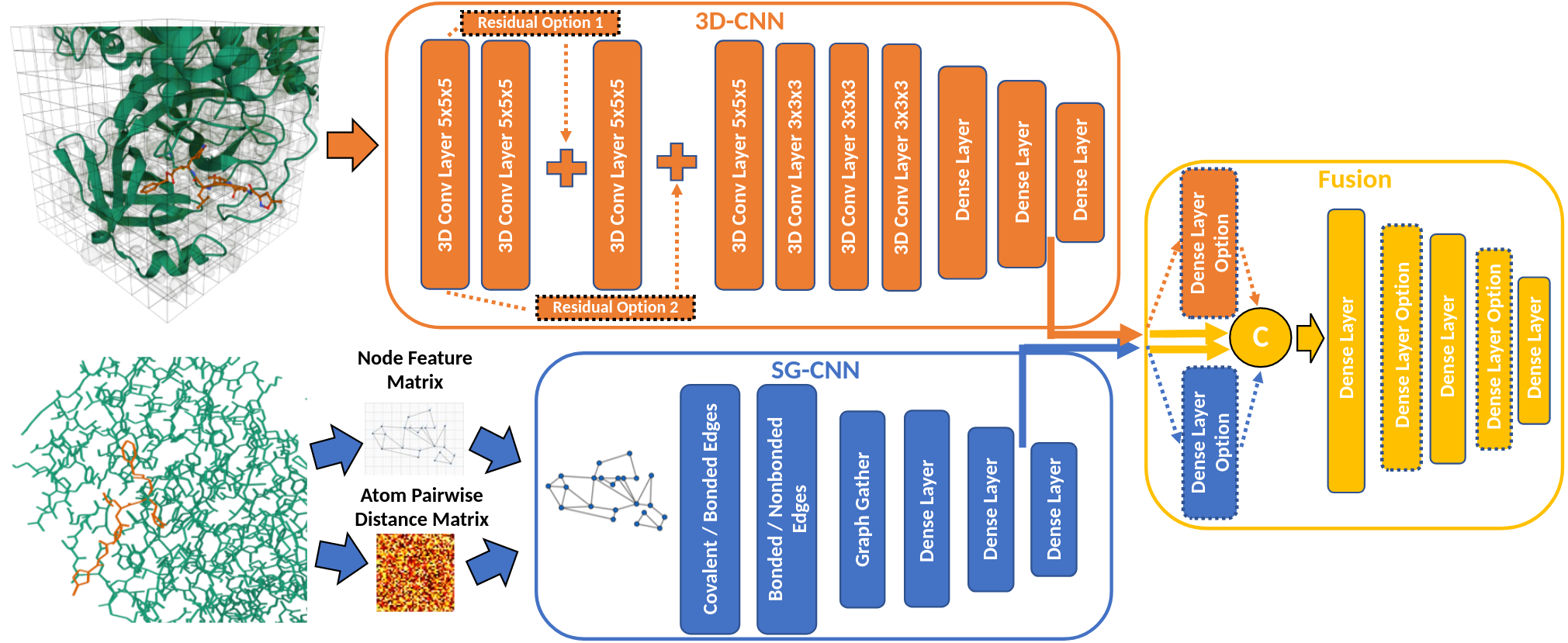}
  \caption{Fusion model architecture with voxelized and spatial graph inputs of COVID-19 M\textsuperscript{pro} (PDB: 6LU7, denoted protease1, green) in complex with an N3 inhibitor (red) [51]. Components of the 3D-CNN (orange), SG-CNN (blue), and Fusion layers (yellow) which were given as options to the hyper-parameter optimization are shown with dashed lines/borders.}
  \label{fig:modelarch}
  \Description{Fusion model architecture}
\end{figure*}
\subsection{Advancing Fusion}
Faced with the imperative task of virtually screening molecules for inhibition of SARS-CoV-2 activity, we sought to apply and improve the fusion deep learning approach. First, we updated the previously trained models with data from the latest version of PDBbind (2019) \cite{wang2005PDBbind}, which provides approximately 4000 more compounds than the 2016 version used to train the original models and, thereby, offers the potential for improved generalization

Our second step looked to improve the 3D-CNN and SG-CNN models through recent developments in the optimization of hyper-parameters. Deep learning models are highly sensitive to a human’s definition of their hyper-parameters – model architecture, loss function, and optimization algorithm, among others \cite{jaderberg2017population}. Automated hyper-parameter optimization was first addressed with parallel searches (grid or random), followed by sequential optimization methods such as Bayesian optimization \cite{srinivas2009gaussian, bergstra2011algorithms, snoek2012practical, hutter2011sequential}. Hyper-parameter optimization algorithms have improved over time in their ability to scale \cite{snoek2015scalable}. Evolutionary Algorithms (EAs) have been shown to further improve the optimization process \cite{jaderberg2017population,castillo2000evolving, husken2000optimization, li2019generalized}. Recently, a leading population-based EA, Population-Based Bandits (PB2) \cite{jaderberg2017population}, was improved by formulating hyper-parameter optimization as a Gaussian Process (GP) bandit optimization of a time-varying function \cite{parker2020provably}.

Fusion modeling in particular is marked by a significant exposure to hyper-parameters.  Both the 3D-CNN and SG-CNN models have their own hyper-parameters which enable them to learn the binding affinity prediction problem optimally in isolation. Additionally, the fusion layers require another set of hyper-parameters necessary to find an optimal non-linear combination of the two models. With this in mind, we saw an opportunity for improving fusion significantly by using a PB2 automated optimization algorithm \cite{parker2020provably} on Lassen, one of the most powerful high-performance computers in the world \cite{lassen}. The libraries used to define the model and optimization architectures are PyTorch \cite{paszke2019pytorch}, Pytorch Geometric \cite{fey2019fast}, and Ray/Ray[Tune] \cite{moritz2018ray}.

Finally, a new formulation of fusion, the Coherent Fusion model, was developed as a potential improvement to the previous Late and Mid-level Fusion models. In both the existing fusion approaches, a 3D-CNN and SG-CNN are individually optimized to minimize the mean squared error (MSE) between their predictions of binding free energy and ground-truth experimental values. The existing Mid-level approach then combines the independently optimized models to form a stronger predictor, by learning the latent space strengths of each model. However, in both Late and Mid-level Fusion, the 3D-CNN and SG-CNN weights are unaltered and remain in the state that was optimal for isolated prediction.
Given the Late and Mid-level Fusion models’ superior performance compared to their isolated components, we hypothesized that fusion might be further improved by coherently backpropagating gradient through both the fusion layers and the separate models. In doing so, the Coherent Fusion model fine-tunes both the 3D-CNN and SG-CNN heads to cooperatively exploit their strengths in a joint optimization. The drawbacks of Coherent Fusion are both an increased hyper-parameter search space and number of trainable parameters. To address this, we developed a parallel, distributed hyper-parameter optimization training architecture. Compared with the models in \cite{jones2020improved}, the combination of these modifications to the concept of Fusion led to significant differences in the hyper-parameters for the 3D-CNN, SG-CNN, and Fusion layers of the model.

\section{Optimization and Evaluation}
\subsection{Data}
PDBbind-2019 is a curated subset of the larger Protein Data Bank (PDB) \cite{wwpdb2019protein}, which is widely used to tune biophysics- and machine learning-based methods \cite{wongetal,jones2020improved, jimenez2018k, feinberg2018potentialnet, lim2019predicting}. The PDBbind data set is comprised of crystal structures arranged into two groups ($general$ and $refined$) based on size (where protein-ligand compounds containing a ligand with molecular weight $>$1000 Daltons (Da) are excluded from the refined set), data quality (where compounds with a measured $IC_{50}$ but no $K_{i}$ or $K_{d}$ measurements are excluded from the refined set), and resolution of the crystal structure ($<$2.5 Angstroms). From the $refined$ set, a third, $core$ set is extracted using a clustering protocol based on protein-sequence similarity. The $core$ set is compiled to represent a valid test for scoring methods by creating a high-quality subset of compounds sufficiently different from the $general$ and $refined$ sets. As such, we use the $core$ set as a primary means for evaluating the Fusion methods considered and comparing against published literature.

In this study, we employ the quintile sub-sampling method from \cite{jones2020improved} to formulate training and validation sets from the PDBbind-2019 $general$ and $refined$ groupings. The sub-sampling is done independently on the $general$ and $refined$ sets and 10\% of the examples from each are withdrawn to form the validation set. Quintile sub-sampling guarantees both the training and validation sets to represent the full range of binding affinity values across PDBbind, where simple random sampling holds the risk of training and validating models on different sub-spaces of affinity values \cite{ellingson2020machine}. The outcome is a training set of 15,631 complexes, a validation set of 1,731 complexes, and the 290 PDBbind $core$ set complexes are held-out for evaluation. Details of pre-processing and feature extraction for the PDBbind data can be found in \cite{jones2020improved}, where here the same tools \cite{pettersen2004ucsf,maier2015ff14sb,jakalian2000fast,o2011open} and sequence of operations are used.

As a means for additional evaluation, we supplement the $core$ set of 290 crystal structures from PDBbind with virtually-docked representations of the complexes. In practice, docking pose data is used for large-scale virtual screening, but is noisy and error prone since the correct ligand pose is not known until a co-complex is crystallized experimentally. Therefore, a scoring function’s performance in the docking space is critical in gauging its robustness to noise and its pragmatic utility.

We leverage the ConveyorLC toolchain \cite{zhang2013message,zhang2014toward,zhang2017comprehensive} to produce all docking complex data, as it is used in our high-throughput virtual screening pipeline. ConveyorLC generates docking poses using the Vina scoring function \cite{trott2010autodock}, then re-scores up to 10 best docking poses using MM/GBSA on a subset of the larger virtual screen; only a subset is re-scored because MM/GBSA is orders of magnitude more computationally expensive than docking.. This sequence of down selecting to limit the search space, accompanied by increasingly complex analyses, is frequently used in drug discovery pipelines, and even molecular dynamics (MD) simulations can be used before finalizing candidates for physical experimentation. The opportunity for machine learning models, like Deep Fusion, is to replace or supplement a more costly stage of a drug discovery pipeline with either improved accuracy or speed.

\begin{table}
\centering
\caption{Hyper-parameters for each model and their range of values considered by the PB2 optimization}

  \begin{tabular}{|c|c|c|c|c|}
  \hline
    Hyper-parameter & 3D-CNN & SG-CNN & Fusion \\ \hline\hline
    Optimizer & \makecell{Adam\\\cite{kingma2014adam}} & \makecell{Adam\\\cite{kingma2014adam}} & \makecell{Adam \cite{kingma2014adam},  \\ AdamW \cite{loshchilov2017decoupled},\\ RMSprop \cite{graves2013generating}, \\Adadelta \cite{duchi2011adaptive}}  \\ \hline
    Activation function & ReLU & ReLU & \makecell{ReLU  \\ LReLU \cite{xu2015empirical},\\ SELU \cite{klambauer2017self}}  \\ \hline
    Batch size & 8,12,24 & 4,8,12,16 & \makecell{1,2,4,5,8,12,\\16,24,28,34,\\38,48,56}  \\ \hline
    Learning rate & \makecell{$1e^{-6}$ - \\ $1e^{-4}$} & \makecell{$2e^{-4}$ -\\ $2e^{-2}$} & \makecell{$1e^{-8}$ - \\ $1e^{-3}$}  \\ \hline
    \makecell{Model-Specific\\Fusion Layers} & N/A & N/A & \textit{T}/\textit{F}  \\ \hline
    Epochs & 0-150 & 0-350 & 0-500  \\ \hline
    Pre-trained & \textit{F} &\textit{F} &  \textit{T}/\textit{F}  \\ \hline
    Batch norm. & \textit{T}/\textit{F} & \textit{F} & \textit{T}/\textit{F}  \\ \hline
    Dropout 1 (early) & 0.25 & 0 & 0 - 0.50  \\ \hline
    Dropout 2 (mid) & 0.125 & 0 & 0  - 0.25  \\ \hline
    Dropout 3 (late) & 0 & 0 & 0 - 0.125  \\ \hline
    \# of Fusion Layers & N/A & N/A & 3,4,5  \\ \hline\hline
   \# of Dense Nodes & \makecell{40,64,88,\\104,128} & N/A & \makecell{8,24,40,64,\\88,104,128}  \\ \hline
    Residual Option 1 & \textit{T}/\textit{F} & N/A & \textit{T}/\textit{F}  \\ \hline
    Residual Option 2 & \textit{T}/\textit{F} & N/A & \textit{T}/\textit{F}  \\ \hline
    \# of Conv. Filters 1 & 32,64,96 & N/A & 32,64,96  \\ \hline
    \# of Conv. Filters 2 & 64,96,128 & N/A & 64,96,128  \\ \hline\hline

  \makecell{Non-covalent /\\Covalent \\ K} & N/A & \makecell{2,3,4,\\5,6,7,8} & \makecell{2,3,4,\\5,6,7,8}  \\ \hline
   \makecell{Non-covalent /\\Covalent \\ Neighbor Threshold} & N/A & 1.2\AA-5.9\AA & 1.2\AA-5.9\AA  \\ \hline
    \makecell{Non-covalent /\\Covalent \\ Gather Width} & N/A & \makecell{8,24,40,64,\\88,104,128} & \makecell{8,24,40,64,\\88,104,128}  \\ \hline
  \end{tabular}
  \label{tab:hyperparams}
\end{table}
\subsection{Training Architecture}
Our approach to train the various individual and fusion models was executed iteratively. Lassen uses the IBM Spectrum LSF Job Scheduler \cite{ibmknowledgecenter}, which necessitates pausing, rescheduling, and resuming training jobs after a maximum run-time. As the hyper-parameter optimization began to converge, the range of hyper-parameter values was adjusted, when possible, to ensure the lower and upper-bounds of the search space were not limiting factors in model performance. The full scope of hyper-parameters and ranges evaluated for each model are provided in Table \ref{tab:hyperparams}, where the ranges are binary (T/F), a list of options, uniformly sampled continuous variables, or not applicable (N/A).

We used the Ray/Ray[Tune] \cite{moritz2018ray} Python library extensively as the foundation for running trials of individual hyper-parameter combinations within the context of a PB2 optimization. Together, Ray and PyTorch provide the ability to accelerate the training process by distributing individual trials across multiple nodes/GPUs. Each of Lassen’s 792 GPU nodes is made up of 44 3.45 GHz Power9 CPU cores, 4 NVIDIA Volta V100 GPUs each with 16GB of memory, and 256 GB of main memory. Depending on the complexity of each model, we distributed individual hyper-parameter configurations between 1 and 12 ranks (1 rank = 1 GPU, 10 CPU cores, 64 GB memory) to expedite the training process. Each rank also utilized 24 data workers running in parallel to pre-load future batches. The combination of distributed model training and parallel data loading was central to the feasibility of an experiment with this size/scope.

The PB2 hyper-parameter optimization was initialized with a quantile fraction ($\lambda\%$) of 50, a time scale ($T$) in Epochs, a perturbation interval ($t_{ready}$) of 100 Epochs, and an objective function ($Q$) of minimum validation set MSE loss \cite{jaderberg2017population,parker2020provably}. The procedure begins with a population of initial, randomly sampled hyper-parameter hypotheses. As every trial reaches the perturbation interval $t_{ready}$, PB2 looks at the model’s performance and determines if it is above or below the quantile fraction ($\lambda\%$). The best performing trials (above $\lambda\%$) continue, while the under-performing trials clone a top-performing configuration (exploiting) and modify it using a parallel GP-bandit optimization (exploring). The training process produces both an optimal model and important information about how the hyper-parameters considered effect performance.
\subsection{Model Architectures}
We draw heavily from the original FAST network architectures in \cite{jones2020improved}, which holds detailed descriptions of pre-processing, feature extraction, voxel grid sizing and atom propagation, which were unaltered. The following focuses on updates to the models and the final optimized hyper-parameter configurations for each component. For brevity, we list only the final optimized hyper-parameter values, where the advantage of PB2 is in its ability to learn a schedule of hyper-parameters to converge in an end-state \cite{parker2020provably}.
 
\subsubsection{Individual Models}

The SG-CNN in this work is structurally unaltered from \cite{jones2020improved}, which uses the PotentialNet \cite{feinberg2018potentialnet} architecture based on Gated Graph Sequence Neural Networks \cite{li2015gated}. The only notable difference is the size of the dense layers were set according to the Non-covalent Gather Width, such that it was sequentially reduced in size by a factor of 1.5 and then 2. A population of 90 SG-CNN trials produced the final model and hyper-parameter configuration given in Table \ref{tab:sgcnn}. 
\begin{table}
\centering
\caption{Final hyper-parameters for the SG-CNN}
  \begin{tabular}{|c|c|}
  \hline
    Hyper-parameter & Value \\ \hline\hline
    Epochs & 213  \\ \hline
    Batch size & 16  \\ \hline
    Learning rate & $2.66e^{-3}$ \\ \hline
  Non-covalent K & 3  \\ \hline
  Covalent K & 6  \\ \hline
  Non-covalent Neighbor Threshold & 5.22\AA  \\ \hline
  Covalent Neighbor Threshold & 2.24\AA  \\ \hline
  Non-covalent Gather Width &  128 \\ \hline
  Covalent Gather Width & 24  \\ \hline
  \end{tabular}
  \label{tab:sgcnn}
\end{table}

\begin{table}
\centering
\caption{Final hyper-parameters for the 3D-CNN}
  \begin{tabular}{|c|c|}
  \hline
    Hyper-parameter & Value \\ \hline\hline
    Epochs & 75  \\ \hline
    Batch size & 12  \\ \hline
    Learning rate & $4.90e^{-5}$ \\ \hline
    Batch normalization & \textit{F}  \\ \hline
    \# of Dense Nodes & 128  \\ \hline
    \# of Conv. Filters 1 & 32  \\ \hline
    \# of Conv. Filters 2 & 64  \\ \hline
    Residual Option 1 & \textit{F}  \\ \hline
    Residual Option 2 & \textit{T}  \\ \hline
  \end{tabular}
  \label{tab:cnn3d}
\end{table}

The 3D-CNN model is slightly modified from the architecture in \cite{jones2020improved}. The model has dropout above the first two dense layers, 2 additional convolutional layers, the filter sizes begin at 5x5x5 and reduce to 3x3x3, the residual options shown in Figure \ref{fig:modelarch} were fed to the hyper-parameter optimization, and similar to the SG-CNN, the second dense layer size was determined by the optimization and then sequentially reduced by a factor of 2. Again, a population of 90 trials was used, the final hyper-parameter values are given in Table \ref{tab:cnn3d}, where the optimization converged to using the second residual connection shown in Figure \ref{fig:modelarch} and 32 to 64 filters for the 5x5x5 and 3x3x3 convolutional layers respectively. With this larger 3D-CNN architecture (deeper than in \cite{jones2020improved}) we found it beneficial to augment the input matrices for the training set by randomly rotating the input data in X, Y, and Z, each with a 10\% probability of occurring. This random rotational augmentation was applied only to the voxelized representation of a compound. While the compound is fundamentally the same, altering its presentation to the model helps to prevent overtraining (\textit{e.g.,} learning rotation-dependent features) and to increase the effective size of the training data set.
\begin{table}[h]
\centering
\caption{Final hyper-parameters for Mid-level Fusion}
  \begin{tabular}{|c|c|}
  \hline
    Hyper-parameter & Value \\ \hline\hline
    Epochs & 64  \\ \hline
    Batch size & 1  \\ \hline
    Learning rate & $4.03e^{-4}$ \\ \hline
    Batch normalization & \textit{F}  \\ \hline
    Optimizer & Adam \cite{kingma2014adam} \\ \hline
    Activation function & SELU \cite{klambauer2017self}  \\ \hline
    Residual Fusion Layers & \textit{T}  \\ \hline
    Dropout Rate 1 (early) & 0.251  \\ \hline
    Dropout Rate 2 (mid) & 0.125  \\ \hline
    Dropout Rate 3 (late) & $\approx$0  \\ \hline
    \# of Fusion Layers & 5  \\ \hline
  \end{tabular}
  \label{tab:mid}
\end{table}
\subsubsection{Late / Mid-level Fusion}
The Late Fusion method was implemented the same as in \cite{jones2020improved}. On the other hand, the optimization led the Mid-level Fusion model to a modified structure. For Mid-level Fusion, every optional layer (dashed lines) in the yellow Fusion block of Figure \ref{fig:modelarch} was turned on. Table \ref{tab:mid} gives the final hyper-parameters for Mid-level Fusion, which are the output of a 180 individual trial population. The other minor differences are a SELU \cite{klambauer2017self} activation was selected over the previous Leaky-ReLU activation \cite{xu2015empirical}, a final batch size of 1, and the usage of light dropout instead of none.
\begin{table}[h]
\centering
\caption{Final hyper-parameters for Coherent Fusion}

  \begin{tabular}{|c|c|}
  \hline
    Hyper-parameter & Value \\ \hline\hline
    Pre-trained & \textit{T}  \\ \hline
    Epochs & 18  \\ \hline
    Batch size & 48  \\ \hline
    Learning rate & $1.08e^{-4}$ \\ \hline
    Batch normalization & \textit{F}  \\ \hline
    Optimizer & Adam \cite{kingma2014adam} \\ \hline
    Activation function & SELU \cite{klambauer2017self}  \\ \hline
    Residual Fusion Layers & \textit{F}  \\ \hline
    Dropout Rate 1 (early) & 0.386  \\ \hline
    Dropout Rate 2 (mid) & 0.247  \\ \hline
    Dropout Rate 3 (late) & 0.055  \\ \hline
    \# of Fusion Layers & 4  \\ \hline
  \end{tabular}
  \label{tab:coherent}
\end{table}
\subsubsection{Coherent Fusion}
In developing the Coherent Fusion model, it was unclear whether the same 3D-CNN and SG-CNN hyper-parameter configurations found to be optimal in isolation would also be ideal for their collaborative prediction. As such, we gave the optimization the option to load the models individually trained for prediction or re-define their structure and train each head from scratch. Using the pre-trained models led to a significant improvement in validation loss. Therefore, Table V gives the final hyper-parameters for the best performing Coherent Fusion model, which loads the weights from the SG-CNN in Table \ref{tab:sgcnn} and 3D-CNN model described in Table \ref{tab:cnn3d}. 

The Coherent Fusion model experiment optimized a population of 270 individual trials to produce a best performer. Interestingly, the Coherent Fusion model converged to exclude the model-specific dense layer options the Mid-level Fusion model uses (Figure \ref{fig:modelarch}) and used a simpler (4 fusion layers) architecture overall. Additionally, the Coherent Fusion used a larger batch size of 48 and significantly stronger dropout. Across the board, the Coherent Fusion model preferred a simpler Fusion architecture with significantly stronger regularization. Our intuition for this phenomenon is that the Coherent Fusion model adjusting a larger set of learned parameters allows for a simpler architecture, faster convergence, and heavier regularization compared to the Mid-level Fusion model, which serves as preliminary evidence of a stronger predictor.
\subsection{Evaluation Results}
\begin{table}
\centering
\caption{Performance of Fusion models on the PDBbind core set crystal structures}

  \begin{tabular}{|c|c|c|c|c|c|}
  \hline
    Model & RMSE & MAE & R$^{2}$ & \makecell{Pearson\\R} & \makecell{Spearman\\R} \\ \hline\hline
    Pafnucy \cite{stepniewska2017pafnucy} & 1.42 & 1.13 & - & 0.78 & -  \\ \hline
    \makecell{Mid-level\\Fusion} & 1.38 & 1.10 & 0.596 & 0.778 & 0.757  \\ \hline
    Late Fusion & 1.33 & 1.07 & 0.623 & 0.813 & 0.805  \\ \hline
    \makecell{Coherent\\Fusion} & 1.30 & \textbf{1.05} & \textbf{0.640} & 0.807 & 0.802  \\ \hline
    KDeep \cite{jimenez2018k} & \textbf{1.27} & - & - & \textbf{0.82} & \textbf{0.82}  \\ \hline

  \end{tabular}
  \label{tab:core}
\end{table}
Over 60,000 Lassen GPU hours were used to optimize the various models. All model iterations and intervals were not run across the same number of nodes, but our training architecture was run at its peak across 66 Lassen nodes capable of over 7,300 TFLOPS using 2904 CPU cores and 264 GPUs to train in parallel.

In Table \ref{tab:core}, the Coherent Fusion model is shown to outperform the Late and Mid-level Fusion methods on the PDBbind \textit{core} set of 290 compounds. While the difference between the Late and Coherent Fusion methods is only 0.03 RMSE, the genetic optimization of Coherent Fusion produced several nearly identical models, which consistently performed better than Late Fusion in all evaluated metrics. Importantly, the Coherent Fusion model converged to a model structure using an automated process that exceeds the performance of the hand-crafted fusion architecture used in the original Mid-level Fusion \cite{jones2020improved}.  Additionally, we provide a comparison to two other deep learning approaches (KDeep \cite{jimenez2018k} and Pafnucy \cite{stepniewska2017pafnucy}) to view Coherent Fusion's performance in a wider scope.
\begin{figure}
  \includegraphics[width=\linewidth]{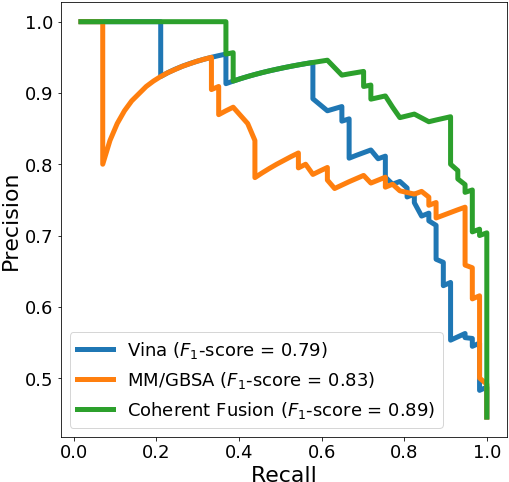}
  \caption{Binary classification of 128 docked complexes from the PDBbind core set, where the positive, ``stronger'' binder class represents 57 compounds with experimental \texorpdfstring{pK\textsubscript{i}}{pK i} or \texorpdfstring{pK\textsubscript{d}}{pK d} > 8 and the negative, ``weaker'' class consists of 71 compounds with \texorpdfstring{pK\textsubscript{i}}{pK i} or \texorpdfstring{pK\textsubscript{d}}{pK d}  < 6.}
  \label{fig:pdbroc}
  \Description{P/R Curves for Vina, MM/GBSA, and Coherent Fusion on the PDBBind core set}
\end{figure}
While the PDBbind \textit{core} set is a standard benchmark for machine learning methods \cite{jones2020improved,jimenez2018k, stepniewska2017pafnucy}, high-throughput virtual screening overwhelmingly relies to docked poses of compounds for drug discovery. To follow suit, we leveraged ConveyorLC \cite{zhang2014toward} to compare Coherent Fusion against physics-based scoring functions in the docking space. The noisier docking data also provides insight into whether the machine learning model was over-trained and how robust it is to noise when scoring more realistic data.

197 compounds from the PDBbind \textit{core} set were successfully evaluated by ConveyorLC with the physics-based Autodock Vina algorithm \cite{trott2010autodock} and MM/GBSA methods for comparison with Coherent Fusion. Each compound was then filtered by $RMSD$, where each of the 197 compounds were checked for a pose with $RMSD < 1 $\r{A} such that a correct pose was found and sufficiently similar to the crystal structure from PDBbind. Using the binding affinity values from PDBbind as ground-truth for the docking poses, Vina achieved a Pearson correlation coefficient of .579, MM/GBSA scored .591, and Coherent Fusion reached .745. To further examine the performance difference between the three methods, binding affinity prediction can be cast as a binary classification problem \cite{jones2020improved}. Figure \ref{fig:pdbroc} shows the results on a subset of the 197 \textit{core} set compounds. Positive and negative classes were created from 57 stronger binders and 71 weaker binders, respectively. Because the set of strong vs. weak docked poses is small (128 total), we elected to compare the different methods using a Precision-Recall Curves and $F_{1}$-scores which give a much more direct picture of how each model is performing than a ROC curve provides.

The nominally small scoring improvements such as, Coherent vs. Late Fusion (0.02 MAE) or Coherent Fusion vs. MM/GBSA (0.06 $F_{1}$-score), have an amplified value in large-scale screening. For example, consider a hypothetical virtual screen evaluating 1 million compounds to subsequently purchase 100 compounds (0.01\%) for experimentation. If the results from classifying the PDBBind core set docking complexes translated to the top 100 of 1 million candidates with a factor of 10 decrease in precision, a virtual screen using MM/GBSA would produce 7 true positives and Coherent Fusion-based screen would produce 9 true positives in < 1/100\textsuperscript{th} of the computational time (Table \ref{tab:speed}). While significant caveats apply, any additional true positive binders are valuable, as in practice, inhibitory compounds are hard to come by and must meet additional pharmacokinetic and safety requirements.
With this analysis, we considered the Coherent Fusion model valuable and validated for use in screening for SARS-CoV-2 inhibitors.

\section{High-Throughput Screening}
\label{sec:high}
Our SARS-CoV-2 effort screened over 500 million compounds against each of the 4 M\textsuperscript{pro} and spike targets, drawing compounds from four public virtual compound libraries \cite{lauetal}. The ZINC database \cite{zinc} was used to create a set of ``world-approved 2018'' drugs from a list of FDA-approved and ``world-not-FDA'' approved drugs. An additional 1.5 million compounds were selected from ChEMBL \cite{gaulton2012chembl}; 18 million compounds were drawn from eMolecules \cite{emolecules}, and the remaining compounds came from Enamine’s list of drug-like compounds estimated to be synthetically feasible \cite{enamine}. 

SMILES strings \cite{smiles} from the eMolecules and Enamine databases and the 2D SDF structures from the ZINC and ChEMBL libraries were downloaded, respectively. Both forms of input were imported to the MOE program \cite{moe} to remove salts and metal-containing ligands, then the protonation states of compounds were set to the dominant form at pH 7. 3D structures of compounds were then generated and energetically minimized. The selected MOE descriptors were calculated and the final structures were exported from MOE as SDF files. These structures were then processed by the ligand preparation in the AMBER tool-chain utilizing antechamber and the GAFF force field \cite{amber}. The 3D SDF files were also converted to PDBQT format for the docking calculations by the Open Babel toolbox \cite{o2011open}. In sum, over 5 billion docking poses were generated and evaluated.

\subsection{Physics-based Screening Pipeline}
As with our comparison between Coherent Fusion and physics-based binding affinity scoring functions, we leveraged the existing ConveyorLC \cite{zhang2014toward} tool chain to search for candidate inhibitors of the two binding sites from spike and active sites for two SARS-CoV-2 protease crystal structures. ConveyorLC is made up of four parallelized programs each designed to handle a specific task in the molecular docking and re-scoring processes. ConveyorLC uses CDT1Receptor to perform protein preparation, CDT2Ligand for ligand preparation, CDT3Docking performs the molecular docking, and finally CDT4mmgbsa handles MM/GBSA re-scoring. Further details on the exact execution, pre-processing, and parameters used in the docking simulations can be found in previous studies \cite{lauetal, zhang2014toward}.

The Vina scoring used in CDT3Docking, operates at approximately one minute per compound per CPU core. On a single Lassen node with 40 CPU cores (each core has 4 hardware threads) using 8 Monte Carlo simulations per compound, Vina is able to dock $\approx$10 docking poses per second. In contrast, a single-point MM/GBSA score takes 10 minutes per docking pose per CPU core. Because of its computational cost, MM/GBSA is often used as a re-scoring function to refine an already filtered set of compounds. \cite{wongetal}. Even on a Lassen node, MM/GBSA is only capable of re-scoring $\approx$0.067 poses per second.
\begin{figure}
  \centering
  \includegraphics[width=\linewidth]{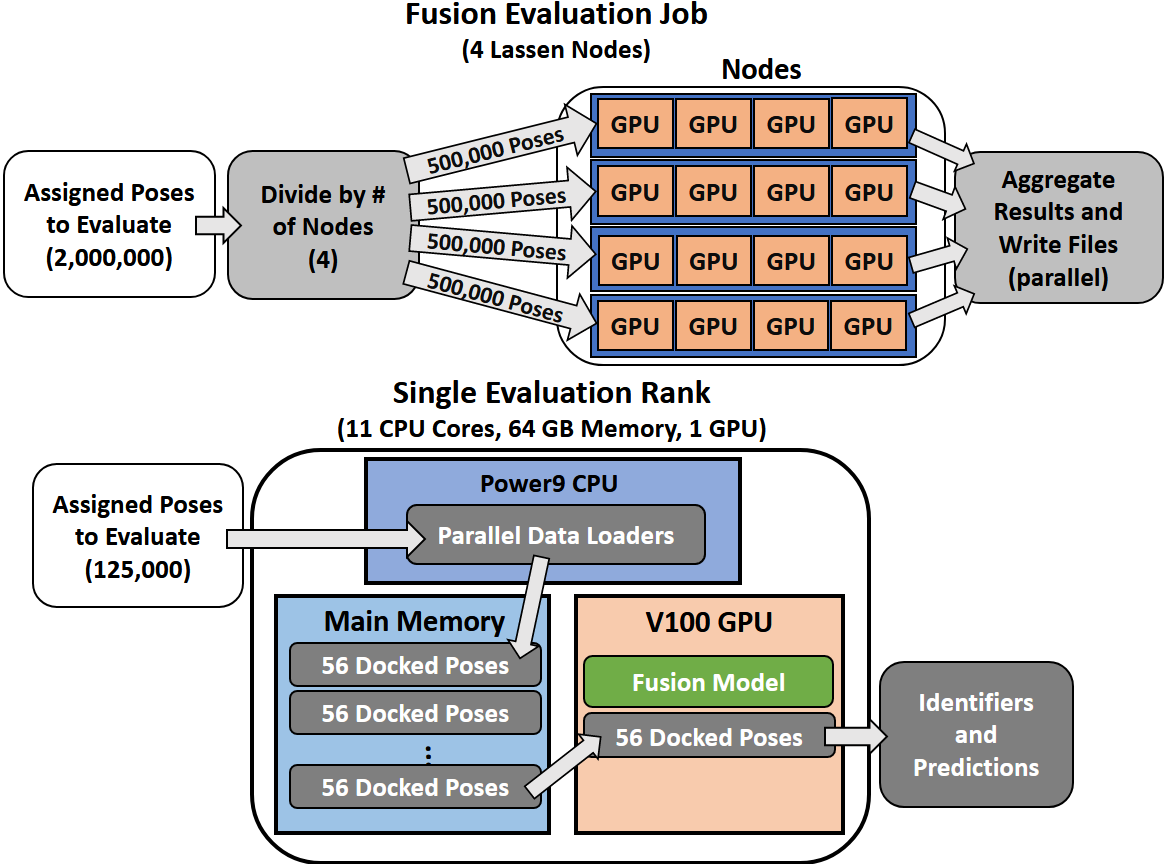}
  \caption{Structure of a single Fusion scoring job (top). A job begins with 2 million poses to score, divides them per node, then each node assigns poses to its ranks and scores. Individual ranks (bottom) take their assigned poses, begin loading batches into memory and feeding them to the GPU for inference. Finally, identifiers and predictions are collected and written in parallel.}
  \label{fig:throughput}
  \Description{Fusion HPC Architecture}
\end{figure}
\subsection{Distributed Fusion Predictions}
In order to screen millions of compounds against SARS-CoV-2, we developed a scalable architecture around the Coherent Fusion model for rapid evaluation (Figure \ref{fig:throughput}). The Coherent Fusion model occupies 1.5 GB GPU memory, which fits on each 16GB NVIDIA Volta V100 GPU. The remainder of the GPU memory is used to simultaneously load 56 individual docked poses into a batch alongside each model. The 4 model instances on each node were given 12 parallel data loaders to accelerate inference. Each model in every job is assigned a subset of compounds to evaluate and its data loaders complete all file reading and pre-processing operations to prepare batches of data in an individual node’s 256 GB of memory, which are subsequently loaded onto the GPU. After evaluating a batch, the screening code unloads the compound, target, and pose identifiers along with the model’s predicted binding affinity. Once a job completes evaluation, the identifiers and predictions are gathered across MPI ranks and distributed across the individual ranks to be written in parallel to HDF5 files. 

In the context of Lassen’s Job Scheduler LSF \cite{ibmknowledgecenter}, we formulated Fusion evaluation jobs as many, individual 4 node processes, each assigned to evaluate an independent set of 2 million poses, which is approximately 200,000 compounds. This format was also a response to our encountering a wide range of errors (bad metadata, node failure, broken pipe errors, etc...), which led to our pipeline being tailored for fault tolerance. With this architecture, when a job fails it has minimal impact on overall throughput (another job takes its place), the reason for failure is easier to pinpoint (log files are smaller and easier to parse), and only a small set of compounds are affected or need to be rescheduled.
\begin{table}[h]
\centering
\caption{Throughput for Fusion prediction single job (2 million poses) and peak performance (125 parallel jobs)}

  \begin{tabular}{|c|c|c|}
  \hline
    Metric & Single Job & Peak\\ \hline\hline
    Avg. Startup & 20 min. & "  \\ \hline
    Avg. Evaluation & 280 min. & "  \\ \hline
    Avg. File Output & 6.5 min. & "   \\ \hline
    Poses per sec. & 108 & 13,594   \\ \hline
    Poses per hour & 338,800  & 48,600,000 \\ \hline
    Compounds per hour & 33,880  & 4,860,000 \\ \hline

  \end{tabular}
  \label{tab:speed}
\end{table}

To create each 4 node job, we relied heavily on Horovod \cite{sergeev2018horovod}, which is based on MPI concepts and uses MPI for cross-node communication. With 4 GPUs on each node, each job is a 16-rank distributed process. Each rank runs a Python script and is given a specific GPU, CPU cores, and memory allocation to execute the evaluation. At the beginning of a job, we simply divide the set of compounds assigned to the job by the number of ranks and assign each rank the subset with its index. When evaluation completes, the ranks use \textit{allgather}  to compile the results and subsequently write out HDF5 files. File output was identified as a bottleneck to the evaluation process early on, which was mitigated by assigning each rank compounds to be written in the same files and directories. The output file format was designed to mirror the output format of ConveyorLC’s CDT3Docking process \cite{zhang2014toward} for interpretation with existing tools and further evaluation of pharmacokinetic and safety properties \cite{lauetal}.

Table \ref{tab:speed} includes a breakdown of how time is spent in an individual job, where the initial 20 minutes are consumed by loading HPC modules, an Anaconda \cite{anaconda} environment, initializing the Horovod ranks, loading an instance of the Fusion model onto each GPU, and pre-loading the initial batches of data for evaluation. The bulk of a job is, as expected, spent in an evaluation period, where batches are loaded, evaluated, and predictions are stored in parallel across all ranks. Finally, once the ranks gather together. The file writing process begins and completes about 6.5 minutes later yielding an average total run-time of approximately 5.1 hours.

With jobs designed for scalability, we regularly ran more than 10 at a time during the SARS-CoV-2 screening effort. However, at set times the majority of Lassen nodes were made available to accelerate evaluation. The peak of which was an allotment of 500 nodes for Fusion screening. The impact of a large number of nodes is clearly seen in Table \ref{tab:speed}, where throughput was increased more than 100 times. Ultimately, during several hours of evaluation at scale, the Coherent Fusion model used more than 14,010 TFLOPS of Lassen’s compute power to screen nearly 5 million compounds per hour. The throughput advantage of Fusion is clear and compared with Vina and MM/GBSA, the Fusion model scoring code provides a 2.7x and 403x speed increase, respectively.

\subsection{Single Job Scalability \& Bottlenecks}
While fault tolerance encouraged the use of many individual jobs each with a small number of nodes, the optimal scale of an individual job was not immediately obvious. Several factors contributed to selecting 4 nodes as optimal including: the 12 hour job run-time limit on Lassen, the startup overhead of a job, the benefit of additional nodes, and the stability of the Python libraries used.

The first parameter explored in development of the Fusion screening capability was the number of poses to evaluate per job. We found it possible to complete up to 5 million scored poses on 4 nodes under the 12 hour Lassen time limit. However, the prevalence of unpredictable errors in the docking data, featurization steps, and inter-node/rank communication, led us to instead assign 2 million poses to each job. While 2 million poses is less efficient (\textit{i.e.,} startup overhead is a larger percentage of each job’s run-time), in practice this decision led to less wasted computational time, as the Fusion scoring code does not write results until it finishes scoring all poses. In future work, efficiency will be improved by creating a separate, parallel process per rank to write results as they are computed, but due to the urgent need for SARS-CoV-2 predictions we elected to mitigate unpredictable errors by narrowing the size of each Fusion job to balance fault-tolerance and computational efficiency. 
\begin{figure}
  \centering
  \includegraphics[width=\linewidth]{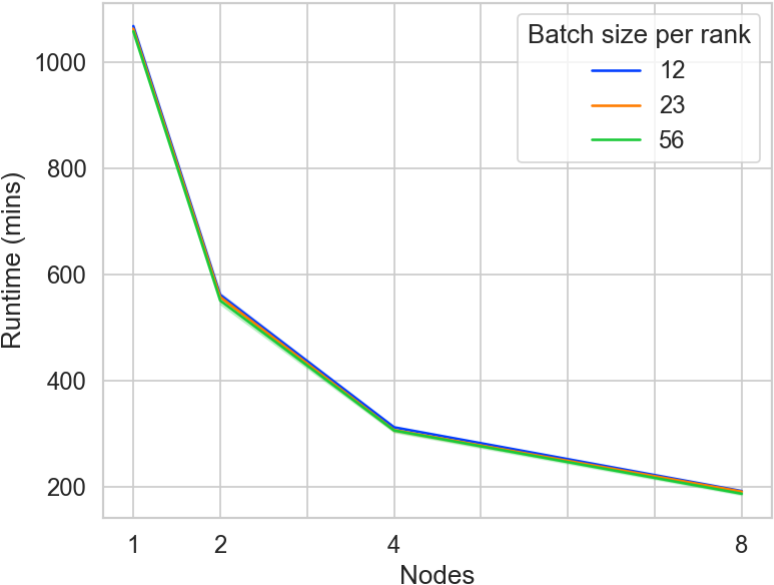}
  \caption{Strong scaling of a single Coherent Fusion job scoring 2 million poses at different batch sizes per rank (12, 23, 56) and number of nodes (1, 2, 4, 8).}
  \label{fig:scaling}
  \Description{Strong scaling for single Coherent Fusion evaluation job.}
\end{figure}

The next two parameters explored were the batch size per rank and number of nodes per job. The batch size per rank effects the number of times data is transferred from CPU->GPU and predictions from GPU->CPU. For the M\textsuperscript{pro} and spike target sites, we found up to 56 poses (each with voxelized and connectivity representations) could consistently fit on the NVIDIA V100 GPU. Figure \ref{fig:scaling} displays the effect three different batch sizes had on a single job. In practice, the performance difference between each batch size was small with a batch size of 56 yielding a $\approx$10 minute run-time advantage over batch size 12.

The Fusion scoring code under-utilized the Lassen GPUs, which led to consistent and relatively small offsets in run-time by batch size. This is due to the computational cost of pre-processing (\textit{e.g.,} file reading and data featurization), which is the most significant bottleneck in the evaluation process. Despite using 12 parallel data loaders per rank, the GPU is intermittently waiting to evaluate more poses. In future work, further optimization of the parallel data loaders will increase GPU utilization and improve throughput, which increases the value of Fusion for large-scale virtual screening.

To determine the optimal number of nodes per job and performance benefit of additional nodes, we evaluated the run-times of Fusion jobs using 1, 2, 4, and 8 nodes per job. Figure \ref{fig:scaling} also displays the performance of each number of nodes where the same set of 10 jobs was run for every point evaluated. The variance between sets is also plotted surrounding each line, but was found to be small (< 5 minutes) and as a result is not clearly visible.

A significant factor in choosing the number of nodes for each job was stability of the Python libraries used by the Fusion model. Inter-node and inter-rank communication errors were increasingly prevalent as the number of nodes/ranks in a job increased and the percentage of failed jobs for each number of nodes was $\approx$2\% for 1 and 2 nodes, $\approx$3\% for 4 nodes, and $\approx$20\% of jobs failed when using 8 nodes. The instability was caused by the specific combination of Horovod \cite{sergeev2018horovod} and PyTorch \cite{paszke2019pytorch} used on the POWER9 architecture, which has since been updated. However, a 20\% job failure rate eliminated 8 nodes per job as a candidate configuration.

The results of our scalability experiments led to a final selection of 2 million poses per job using 4 nodes. This was the result of several different factors and adjustments including the observation that when using 500 Lassen nodes at the same time, the LSF scheduler encountered problems simultaneously running 250 2-node jobs, which was solved by using 125 4-node Fusion jobs instead.

\section{SARS-CoV-2 Results}
The high-throughput virtual drug screening pipeline described in Section \ref{sec:high} produced several computational results for each compound screened. The Fusion model’s binding-affinity prediction was one of the three energy calculations (Vina, MM/GBSA, Fusion), which were used as a component of a hand-tailored cost function designed to filter which compounds to purchase for experimental evaluation and which were less likely to be successful. Full details of the ranking and reasoning may be found in \cite{lauetal} and computational predictions are made available at \url{https://url-excluded} \cite{dataportal}. Virtual screening output on the computational side fed directly into an experimental process to physically interrogate candidate molecules.

\subsection{Experimental Validation}
Experimental testing of the candidate binders which were screened and purchased to target M\textsuperscript{pro} used a fluorescence resonance energy transfer (FRET) based activity assay or a SDS-PAGE gel protein cleavage assay. After assay optimization, additional screens were run in order to down select compounds. For example, compounds from the ZINC database \cite{zinc} were down selected to an additional testing of 19 compounds, which yielded 4 candidates inhibiting the activity of M\textsuperscript{pro} at 100 micro-Molar (\textmu M) concentrations. The four identified compounds include: candestartan cilexetil, FAD disodium, tigecycline, and tetracycline \cite{lauetal}.

On the other hand, compounds predicted to inhibit the SARS-CoV-2 spike protein were screened by both a pseudo-typed virus assay and a biolayer inferometry competitive assay (BLI). Here the candidate compounds are being evaluated for their ability to inhibit ACE2-spike binding and in parallel, the spike binding candidates were screened using a cell-based infection assay at 10\textmu M. Further details of the experimental design, assays, results, and discussion can be found in \cite{lauetal}. 
\begin{figure}
  \centering
  \includegraphics[width=\linewidth]{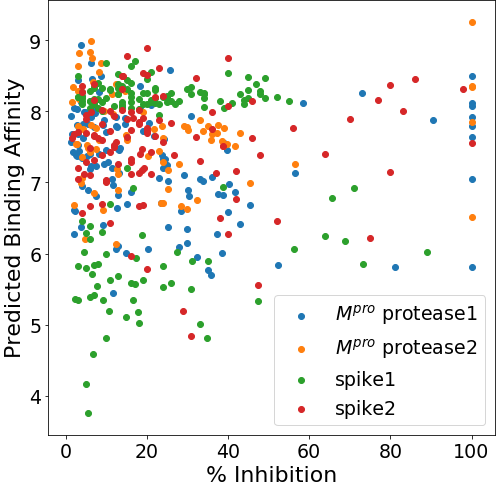}
  \caption{Coherent Fusion predicted binding affinity vs. experimental percentage of inhibition at 100\textmu M for 130 compounds against M\textsuperscript{pro} protease1 (blue) and 81 compounds against M\textsuperscript{pro} protease2 (orange). The spike assays were evaluated at 10\textmu M and include 151 compounds against spike1 (green) and 113 compounds against spike2 (red). Compounds which exhibited $\leq$1\% inhibition (no experimental binding activity) are excluded.}
  \label{fig:predvsactual}
  \Description{Scatter plot of Fusion predictions vs experimental percent inhibition.}
\end{figure}
\subsection{Connecting Predictions and Experimental Results}
Given the ground-truth experimental values for physically tested compounds, we’re enabled to retrospectively evaluate the accuracy of each computational approach which generated a prediction. Some of the obvious questions to ask are: ``Which method was most correlated with the experimental results?", ``Are the most accurate scoring functions the same for all four M\textsuperscript{pro} and spike targets?", and ``Which of the scoring methods is most accurate for the strongest experimental inhibitors?".
\begin{figure*}
  \centering
  \includegraphics[width=\textwidth]{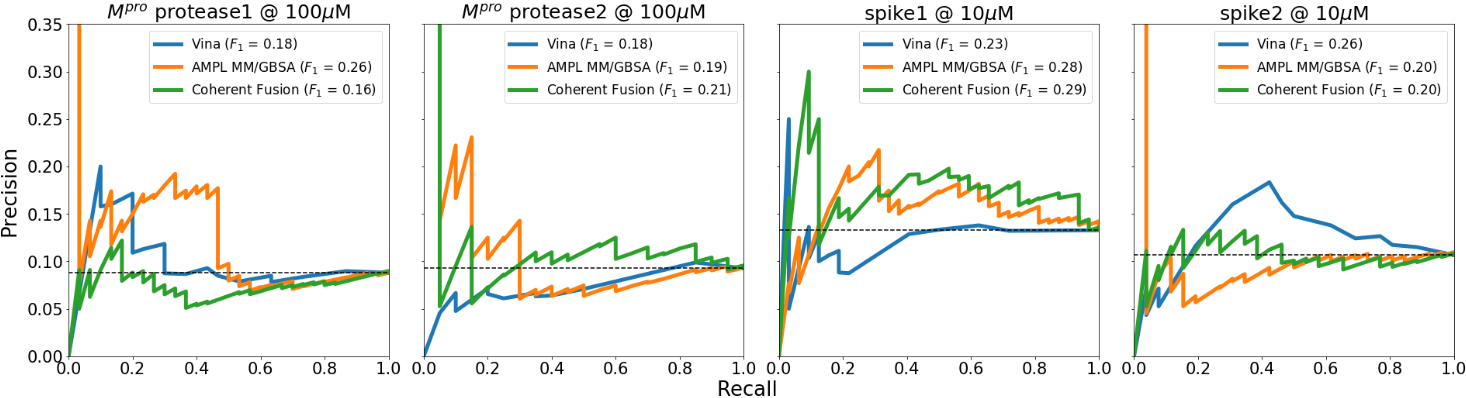}
  \caption{Precision/Recall Curves and F1-scores by SARS-CoV-2 protein target at 33\% experimental inhibition. M\textsuperscript{pro} protease1 (far left) shows results for 30 positive and 311 negative binders. M\textsuperscript{pro} protease2 (middle left) includes 20 positive and 196 negative binders. The spike1 site (middle right) includes 32 positive and 209 negative binders. Finally, spike2 (far right) includes 26 positive and 218 negative binders. The black horizontal dashed line indicates the performance of a random classifier.}
  \Description{P/R Curves for Vina, MM/GBSA, and Coherent Fusion predictions vs. experimental percent inhibition.}
  \label{fig:prcurves}
\end{figure*}
Each experimentally prosecuted compound can be traced back to its virtually docked poses for either the two M\textsuperscript{pro} or two spike binding targets. This means each scoring method may have predicted binding affinity values for up to 40 poses per compound (10 poses maximum per binding site). While in the computational domain we have predicted binding affinity values, the output from the M\textsuperscript{pro} and spike assays is a percentage of inhibition normalized between 0 and 100\%. These values are produced at a given concentration of the candidate drug (100\textmu M for M\textsuperscript{pro} targets and 10\textmu M for spike targets), which gives context to the overall strength of a binder.

For each scoring method (Vina, MM/GBSA, Coherent Fusion) the results per compound were aggregated and represented by the strongest prediction across all poses for each binding site (maximum for Coherent Fusion, minimum for Vina and MM/GBSA). Because of the computational cost of MM/GBSA, we instead use the ATOM Modeling PipeLine (AMPL) MM/GBSA predicted MM/GBSA values, which have been shown to be highly correlated with actual MM/GBSA calculations and were trained to predict MM/GBSA scores on each specific target \cite{mcloughlin}.

\subsection{Analyzing Computational Predictions}
Following this aggregation, the output is a single prediction per compound tied to a single percent inhibition. Each method can then be viewed as a scatter plot comparing predicted vs. actual experimental results as in Figure \ref{fig:predvsactual}. We computed Pearson and Spearman correlation coefficients for each method across all experimentally tested M\textsuperscript{pro} and spike compounds. However, most experimentally tested compounds are negatives ($\leq1\%$ inhibition), which gives correlation coefficients near 0 for each of the three evaluated methods (table excluded for brevity). In an attempt to focus our analysis on the relative strengths and weaknesses of the different scoring methods and not the difficulty of the overall binding affinity prediction problem, we computed correlation coefficients for each method on the subset of compounds for which any experimental binding (>1\%) was observed.

\begin{table}
\centering
\caption{Correlation of predicted binding and percent inhibition on compounds with $>1\%$ inhibition}

  \begin{tabular}{|c|c|c|c|}
  \hline
    Method & Target/Site & \makecell{Pearson\\R} & \makecell{Spearman\\R} \\ \hline\hline
    Vina & M\textsuperscript{pro}/protease1 & 0.03 & -0.08  \\ \hline
    AMPL MM/GBSA & M\textsuperscript{pro}/protease1 & \textbf{0.08} & \textbf{0.01}  \\ \hline
    Coherent Fusion & M\textsuperscript{pro}/protease1 & -0.06 & -0.04  \\ \hline\hline
    Vina & M\textsuperscript{pro}/protease2 & -0.08 & -0.14 \\ \hline
    AMPL MM/GBSA & M\textsuperscript{pro}/protease2 & -0.05 & -0.07   \\ \hline
    Coherent Fusion & M\textsuperscript{pro}/protease2 & \textbf{0.04} & \textbf{0.04} \\ \hline\hline
    Vina & spike/spike1 & -0.02 & 0.06 \\ \hline
    AMPL MM/GBSA & spike/spike1 & 0.15 & 0.22   \\ \hline
    Coherent Fusion & spike/spike1 & \textbf{0.22} & \textbf{0.30} \\ \hline\hline
    Vina & spike/spike2 & \textbf{0.13} & \textbf{0.27} \\ \hline
    AMPL MM/GBSA & spike/spike2 & -0.02 & -0.05   \\ \hline
    Coherent Fusion & spike/spike2 & -0.02 & -0.01 \\ \hline

  \end{tabular}
  \label{tab:corr}
\end{table}

Table \ref{tab:corr} shows the correlations for all methods where the absolute value of the Vina and MM/GBSA scores are used. AMPL MM/GBSA gives the best correlation for the protease1 target, Coherent Fusion for the protease2 and spike1 targets, and Vina scores the spike2 binding site best. However, across the board, it is clear that even when limiting the analysis to >1\% inhibition, the correlations for each method remain low and the interpretation of near-zero correlation coefficients is unavailing.

While removing all the non-inhibitors gives a glimpse into which methods are most correlated with the SARS-CoV-2 binders, it also removes the context of those predictions. That is, the overall prediction strength for each method is somewhat obscured as the range of each method’s prediction values is limited to its minimum and maximum prediction in the smaller set of SARS-CoV-2 binders.
\begin{figure*}
  \centering
  \includegraphics[width=\textwidth]{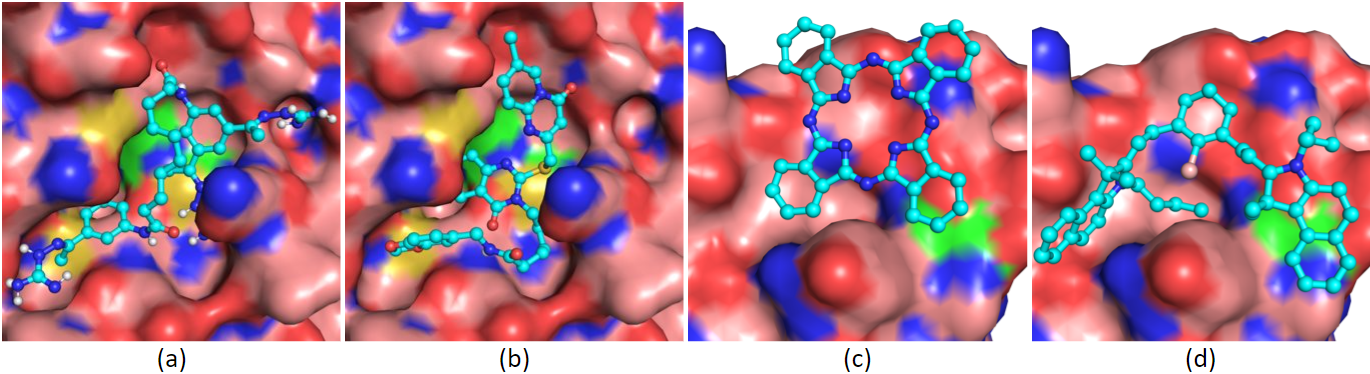}
  \caption{Four compounds from eMolecules \cite{emolecules} in complex with the M\textsuperscript{pro}/protease1 (a, b) and spike/spike1 targets (c, d), where residues His41 and Cys145 in M\textsuperscript{pro} and residue 501 in the spike RBD protein are green. Panels (a) and (b) show Compound IDs 76051337 and 24424612, respectively, which both had 100\% inhibition at 100\textmu M in the M\textsuperscript{pro} assay. Panels (c) and (d) show Compound IDs 18594404 and 313102183, respectively, which reached 100\% and 98\% inhibition at 10\textmu M in the spike assay.}
  \Description{Four top ligands from the eMolecules \cite{emolecules} library in complex with the M\textsuperscript{pro}/protease1 (Panels a,b) and spike/spike1 target sites (Panels c, d).}
  \label{fig:compounds}
\end{figure*}
With this in mind, we sought to answer the question of which scoring methods were most accurate for the strongest experimental inhibitors by including non-binding compounds and casting the prediction problem as a binary classification of compounds with >33\% inhibition (positive class) and compounds with $\leq33\%$ inhibition (negative class). A threshold of 33\% was chosen to avoid severe class imbalances caused by higher thresholds. This problem formulation is similar to that in Figure \ref{fig:pdbroc} where we set a threshold to separate stronger binders from weaker binders. This approach is applicable in practice, as virtual screens eventually down select to a small set of candidates for final analysis, purchase, and experimental testing.

Figure \ref{fig:prcurves} displays Precision/Recall Curves and $F_{1}$-scores using a threshold of 33\% to separate the positive and negative binding examples for each target. The y-axis of each P/R curve is limited to 0.35 to observe different model behaviors. Although the curves for each target appear different from the P/R plots in Figure \ref{fig:pdbroc}, they provide information about how each model performs in practice on a noisier experimental screen, where the cutoff separating active from inactive molecules is less clear.

While the P/R Curves in Figure \ref{fig:prcurves} give low $F_{1}$-scores and precisions, two important observations arise. First, for each of the four binding sites, we computed Cohen’s Kappa statistic ($\kappa$) to compare each model with a random classifier \cite{cohen1960coefficient, landis1977measurement}. Equation \ref{eq:2} gives the equation for computing $\kappa$;
\begin{equation}
\kappa = \frac{\rho_{o} - \rho_{e}}{1 - \rho_{e}}\label{eq:2}
\end{equation}
where $\rho_{o}$ is the observed accuracy and $\rho_{e}$ is the expected accuracy. A random classifier achieves a $\kappa=0$ by guessing according to the frequency of the positive and negative classes. Every model on each target was measured to achieve a $\kappa$ > 0 (with the exception of Vina on spike1), indicating the models are consistently better than random across the targets. In the context of P/R Curves, a random classifier’s performance is expressed as a horizontal line (Figure \ref{fig:prcurves}) with constant expected precision equal to the proportion of positive examples in the data set. 

In practice, the three binding affinity prediction models served as input to a hand-crafted cost function \cite{lauetal}, to select compounds for experimentation. The outcome of which produced 9 distinct compounds inhibiting 100\% of M\textsuperscript{pro} activity and 108 total compounds with $\geq$33\% inhibition. Using the 33\% inhibition threshold, 108 of 1042 (10.4\%) experimentally tested compounds inhibit activity, indicating the models have significant predictive power. In that, while 500+ million candidates were evaluated, only a fraction (2.1$e^{-6}$\%) of the best candidates were tested experimentally, yet the models successfully yielded a 10.4\% hit rate.

Figure \ref{fig:compounds} presents four top compounds bound to two of the four sites/configurations used in the virtual screen (M\textsuperscript{pro}/protease1 and spike/spike1). The M\textsuperscript{pro}/protease1 compounds reached 100\% inhibition and had Coherent Fusion predicted binding affinities of 8.5 (Fig. \ref{fig:compounds}a) and 8.1 (Fig. \ref{fig:compounds}b). The spike RBD compounds evaluated at 10\textmu M had predicted affinities and experimental inhibitions at 7.6/100\% (Fig. \ref{fig:compounds}c) and 8.3/98\% (Fig. \ref{fig:compounds}d).

For the M\textsuperscript{pro} protease1 binding site, AMPL MM/GBSA is the best predictor of experimental binding. Additionally, between the two M\textsuperscript{pro} binding sites, protease1 and protease2, the AMPL MM/GBSA predictions on protease1 conjointly achieve a higher $F_{1}$-score and Pearson correlation coefficient than all other methods. In fact, for each of the four targets, not only do the best $>$33\% $F_{1}$-scores match with the best $>$1\% Pearson/Spearman correlations, but the relative differences across target binding sites are also in agreement.

Coherent Fusion reaches the maximal $F_{1}$ and correlation coefficients for the M\textsuperscript{pro} protease2 and spike1 binding sites. Across all model / binding site combinations, Coherent Fusion’s performance on the spike1 protein is the top performer, though followed closely by AMPL MM/GBSA on the same target site. This is unexpected, as the protease targets are much larger and nominally thought to provide better opportunity for drug-like compounds to bind.

However, target specific strengths are not unexpected \cite{wongetal}. Especially where the M\textsuperscript{pro} sites are large protein pockets and the spike targets are much smaller. Interestingly, the maximal >1\% correlations and >33\% $F_{1}$-scores across all targets favor the spike proteins. The differing concentrations between the M\textsuperscript{pro} and spike experiments is important to note, as M\textsuperscript{pro} binders were evaluated at 100\textmu M, which is a higher drug concentration and, therefore, allows for weaker binders to exhibit higher observed percentages of inhibition.

\section{Conclusion and Future Work}
Deep Fusion was improved by coherent backpropagation and distributed, genetic hyper-parameter optimization. The optimization automatically produced an version of the Coherent Fusion model, which was shown to exceed the performance of hand-crafted Fusion model variants and alternative machine learning methods on crystal structures from the PDBbind \textit{core} set benchmark. In evaluating noisier docked poses of the same test set, Coherent Fusion also achieved an improvement in correlation and $F_{1}$-score relative to the physics-based Vina and MM/GBSA methods. 

We utilized Coherent Fusion to screen over 500 million compounds across four SARS-CoV-2 binding sites. This was achieved by designing a high-throughput distributed architecture for fault-tolerant scoring at scale. Using parallel data loaders and Lassen’s 4 GPUs per node, the Coherent Fusion model is currently capable of screening $\approx$30 poses per second per node. Coherent Fusion was used as input to a weighted cost function across binding affinity and other scoring models in order to sub-select compounds for experimentation.  

Among the nearly 1000 compounds tested experimentally, several inhibited activity of M\textsuperscript{pro} and spike. In analyzing which binding affinity methods were most correlated with the experimental results, we found the optimal performer to vary by target. However, two of the four targeted binding sites were best predicted by Coherent Fusion both in terms of >1\% inhibiting correlation with the experimental results and >33\% inhibiting binary classification.  Although overall predictive power of all of the scoring functions is limited, any opportunity for computational enrichment of strong binders is needed to alleviate the otherwise prohibitive time and cost of the physical experimental screens.

In future work, we aim to use our baseline Coherent Fusion model from this work to fine tune and predict for specific protein target types and binding sites. We believe introducing target specificity to the models and thereby reducing the scope of the binding affinity prediction problem will increase the value of relative differences in the model’s binding affinity predictions. Our high-throughput architecture can also be accelerated, as the GPUs on each Lassen node were observed to be under-utilized. Increased numbers of parallel data loaders were observed to decrease the overall stability of each individual evaluation job indicating some refactoring is necessary.   

\begin{acks}
The authors gratefully acknowledge Professor Xinquan Wang (Tsinghua University) for providing early access to his crystal structure of the ACE2-RBD complex. The authors gratefully acknowledge extensive computer time provided by Livermore Computing. Part of this research was supported by the DOE Office of Science through the National Virtual Biotechnology Laboratory, a consortium of DOE national laboratories focused on response to COVID-19, with funding provided by the Coronavirus CARES Act. The authors thank Lawrence Livermore National Laboratory for funding Laboratory Directed Research and Development projects 20- ERD-065 and 20-ERD-062. Part of this research was also supported by the American Heart Association under CRADA TC02274-4 and the National Nuclear Security Administration through the Accelerating Therapeutics for Opportunities in Medicine (ATOM) Consortium under CRADA TC02349 . This work was funded in part by DTRA under award HDTRA1036045. Sandia National Laboratories is a multimission laboratory managed and operated by National Technology \& Engineering Solutions of Sandia, LLC, a wholly owned subsidiary of Honeywell International Inc., for the U.S. Department of Energy’s National Nuclear Security Administration under contract DE-NA0 0 03525. All work performed at Lawrence Livermore National Laboratory is performed under the auspices of the U.S. Department of Energy under Contract DE-AC52-07NA27344. Any subjective views or opinions that might be expressed in the paper do not necessarily represent the views of the U.S. Department of Energy or the United States Government.
\end{acks}

\bibliographystyle{ACM-Reference-Format}
\bibliography{sample-base}


\end{document}